\documentclass[runningheads]{llncs}
\usepackage[T1]{fontenc}
\usepackage{graphicx}
\usepackage{cite}
\usepackage{booktabs}
\usepackage{xcolor}
\usepackage{multirow}
\usepackage{amsmath}
\usepackage{tabularx}
\usepackage{placeins}
\usepackage{xcolor}
\usepackage{amssymb}
\usepackage[colorlinks, linkcolor=red, anchorcolor=blue, citecolor=blue]{hyperref} 
\definecolor{SeaGreen4}{RGB}{0,205,102} 
\definecolor{SlateBlue}{RGB}{106,90,205} 
\definecolor{DarkRed}{RGB}{178,34,34} 
\begin{document}

\title{SNN-PAR: Energy Efficient Pedestrian Attribute Recognition via Spiking Neural Networks
\thanks{Corresponding author: Xiao Wang, Email: (\email{xiaowang@ahu.edu.cn})}
}


\author{
Haiyang Wang \inst{1} \and 
Qian Zhu \inst{1} \and 
Mowen She \inst{1} \and 
Yabo Li \inst{1} \and 
Haoyu Song \inst{1} \and \\ 
Minghe Xu \inst{2} \and 
Xiao Wang* \inst{1} 
}      

\authorrunning{Haiyang Wang et al.}

\institute{
School of Computer Science and Technology, Anhui University, Hefei, China \and  
Faculty of Data Science, City University of Macau, Macau, China  
}

\maketitle              

\begin{abstract}
Artificial neural network based Pedestrian Attribute Recognition (PAR) has been widely studied in recent years, despite many progresses, however, the energy consumption is still high. To address this issue, in this paper, we propose a Spiking Neural Network (SNN) based framework for energy-efficient attribute recognition. Specifically, we first adopt a spiking tokenizer module to transform the given pedestrian image into spiking feature representations. Then, the output will be fed into the spiking Transformer backbone networks for energy-efficient feature extraction. We feed the enhanced spiking features into a set of feed-forward networks for pedestrian attribute recognition. In addition to the widely used binary cross-entropy loss function, we also exploit knowledge distillation from the artificial neural network to the spiking Transformer network for more accurate attribute recognition. Extensive experiments on three widely used PAR benchmark datasets fully validated the effectiveness of our proposed SNN-PAR framework. The source code of this paper will be released on \url{https://github.com/Event-AHU/OpenPAR}.  
\end{abstract}

\section{Introduction} 

Pedestrian Attribute Recognition (PAR)~\cite{wang2022PARsurvey, cheng2022VTB } targets describing the appearance cues of humans from an attribute set, like gender, age, hair style, wearing, etc. It is a widely studied research problem due to its important role in human-related tasks, such as person re-identification~\cite{lin2019improving}, detection and tracking~\cite{li2024attmot}, text-based retrieval~\cite{huang2024attPersonRetrieval}. With the development of deep neural networks, the research on the PAR has been widely exploited using different neural networks~\cite{cheng2022VTB, deepmar, wang2017JRL, wang2024empiricalmamba} and training strategies~\cite{fan2023parformer, lu2023oagcn}. However, the challenging factors still influence the final results significantly including illumination, background clutters, and motion blur. 

With the aforementioned issues in mind, we first review existing PAR models and find that the mainstream neural networks like the CNN~\cite{he2016resnet}, RNN~\cite{chung2014gru}, Transformer~\cite{vaswani2017Transformer, DosovitskiyViT, wang2023MMPTMs, zhao2023transformerVLT, wang2021TNL2K} are widely utilized for this problem. To be specific, Wang et al.~\cite{wang2017RNNPAR} propose JRL, a novel framework for pedestrian attribute recognition that employs LSTM~\cite{hochreiter1997lstm} to jointly learn attribute context and correlations in a recurrent manner. Transformer networks, initially introduced for natural language processing tasks, they gain adoption within the computer vision community because of their remarkable performance~\cite{vaswani2017Transformer, DosovitskiyViT, wang2023MMPTMs, zhao2023transformerVLT, wang2021TNL2K}. Several studies explore the use of Transformers in the PAR domain to capture global contextual information~\cite{tang2022drformer, cheng2022VTB}. For instance, DRFormer~\cite{tang2022drformer} models long-range relationships between regions and attributes, while VTB~\cite{cheng2022VTB} integrates image and language information to achieve more accurate attribute recognition.
Although these works improve the PAR performance significantly, however, the inference cost is still high in the testing stage.

Recently, Spiking Neural Networks (SNN) have drawn more and more attention due to their advantages of lower energy consumption and bio-inspired network design. Various spiking neurons (LIF~\cite{eshraghian2023trainingspikingneuralnetworks}, ALIF~\cite{sung2020training}) are developed to replace the artificial neurons (e.g., ReLU) in MLP, CNN, or Transformer networks, thus, leading to spiking versions of these models. SNN has been widely used in object detection~\cite{burelo2020spikingneuralnetworksnn}, recognition~\cite{kabilan2021neuromorphic}, tracking~\cite{xiang2024spiking, bing2019end}, image enhancement and reconstruction~\cite{duwek2021image, zhu2022event}, but few efforts are conducted on the pedestrian attribute recognition. Consequently, it makes sense to ask the subsequent question "\textit{How can we design an energy-efficient spiking backbone network for pedestrian attribute recognition?}"

In this work, we propose the first spiking Transformer networks for the PAR task, as shown in Fig.~\ref{fig:framework}. Given the pedestrian image, we first adopt a spiking tokenizer module to get the spiking features, which contain Conv-BN-Multistep LIF-MaxPooling-Conv-BN layers. Here, the Conv and BN are short for Convolutional and Batch Normalization layers, respectively. The Multistep LIF spiking neuron is used as the activation function. The output features will be fed into a spiking Transformer block, each block contains a core self-attention operation and residual connections. We feed the spiking features into a set of FFN (Feed Forward Networks) for attribute prediction. The BCE (Binary Cross-Entropy) loss function is used for the optimization of the whole SNN-PAR framework. To improve the final recognition performance, we further introduce the knowledge distillation strategy to guide the optimization of the SNN-PAR network. In our implementation, the VTB~\cite{cheng2022VTB} is selected as the teacher network for knowledge distillation. 
We conducted extensive experiments on three PAR benchmark datasets and these results fully validated the effectiveness of our SNN-PAR framework for pedestrian attribute recognition.

In conclusion, we highlight the contributions of this paper in the following three areas: 

1). We propose an energy-efficient spiking Transformer network for pedestrian attribute recognition, termed SNN-PAR. To the best of our knowledge, it is the first work that exploit the SNN for the PAR task. 

2). To enhance the performance of SNN-PAR further, we adopt knowledge distillation from the artificial neural networks to guide the learning of spiking Transformer networks.  

3). Comprehensive experiments carried out on three publicly available datasets show that our proposed SNN-PAR model is effective for the PAR task.

\section{Related Works} 

\subsection{Spiking Neural Networks} 
Spiking Neural Networks (SNNs), hailed as the third generation of neural networks, aim to emulate the complex information processing mechanisms of the human brain. Due to their low power consumption characteristics, an increasing number of studies~\cite{shen2022backpropagation, wei2024event, wu2021progressive, zeng2023braincog} and innovative Spiking Transformer architectures ~\cite{yao2024spike, zhou2022spikformer} have emerged. Federico et al.~\cite{paredes2019unsupervised} propose a hierarchical spiking architecture for optical flow estimation, leveraging selectivity to local and global motion through Spike-Timing-Dependent Plasticity (STDP)~\cite{caporale2008spike}. Zhou et al.~\cite{zhou2021temporal} utilize non-leaky Integrate-and-Fire (IF) neurons with single-spike temporal coding to train deep SNNs. Additionally, Zhou et al. combine spiking neurons with Transformer networks to introduce the Spikformer~\cite{zhou2023spikingformer} for recognition tasks. Fang et al.~\cite{fang2021incorporating} introduces an innovative method for simultaneously learning synaptic weights and membrane time constants in SNNs. SNNs are developed using two primary training techniques: conversion from ANNs and direct training. The conversion method employs pulse frequencies to emulate ReLU activation, which aids in transforming pre-trained ANNs into SNNs~\cite{cao2015spiking, zeng2023braincog}. Conversely, the direct training method applies alternative gradient approaches to optimize SNNs directly~\cite{zeng2023braincog}, allowing these networks to be trained on diverse datasets and achieve competitive results within a short number of time steps. This methodology has resulted in a broad range of applications for SNNs in visual tasks, including essential areas such as image recognition, natural language processing, and robotic control ~\cite{national1931proceedings, zheng2021going, zhu2023spikegpt}. This underscores the capability of SNNs to perform computationally demanding tasks efficiently, presenting a sustainable option compared to traditional ANNs.In this study, we investigate pedestrian attribute recognition using SNNs and adopt the direct training strategy to construct our model.

\subsection{Pedestrian Attribute Recognition} 
Pedestrian attribute recognition~\cite{wang2022PARsurvey} uses predefined images to predict pedestrian attributes through various model architectures, including CNNs~\cite{abdulnabi2015multi, zhang2014panda}, RNNs~\cite{wang2016cnn, tian2015pedestrian}, GNNs~\cite{li2019visual, park2017attribute}, and Transformers~\cite{fan2023parformer, cheng2022VTB}. Early works rely on CNNs for attribute analysis. Specifically, Abdulnabi et al.~\cite{abdulnabi2015multi} use CNNs for attribute analysis. They introduce a multi-task learning strategy that employs several CNNs to acquire attribute-specific features, allowing for knowledge sharing between the networks. Zhang et al. introduce the PANDA~\cite{zhang2014panda}, which combines part-aware models with CNN-based attribute classification.

RNNs are employed to model sequential dependencies in attributes. For example, Wang et al.~\cite{wang2016cnn} Use Long Short-Term Memory (LSTM) to develop robust semantic connections among labels in the context of pedestrian attribute recognition. By integrating labels that have been predicted earlier, the visual features can flexibly adjust to accommodate the subsequent labels. Zhao et al. propose the GRL~\cite{tian2015pedestrian} to model attribute dependencies, addressing both intra-group exclusions and inter-group associations. Graph Neural Networks (GNNs) are introduced to model semantic relations among attributes. VC-GCN~\cite{li2019visual} and A-AOG~\cite{park2017attribute}
depict attribute correlations using graphical models. Li et al.~\cite{li2019visual} approach pedestrian attribute recognition as a sequence prediction task, leveraging GNNs to represent spatial and semantic relationships. Recently, Transformer models, leveraging self-attention mechanisms, have gained prominence in the PAR task. Numerous works in PAR have also been developed utilizing the Transformer network. For instance, Fan et al.~\cite{fan2023parformer}present a model called PARformer, which extracts features using Transformers instead of CNNs, effectively merging both global and local perspectives. VTB~\cite{cheng2022VTB} integrating an additional text encoder to enhance pedestrian attribute recognition. 
Different from these works, this paper first exploits energy-efficient PAR using spiking Transformer networks.

\subsection{Knowledge Distillation}

Knowledge Distillation is a technique for model compression that facilitates a smaller student model in learning from a larger teacher model. The student acquires knowledge by imitating various aspects of the teacher, such as its intermediate features~\cite{romero2014fitnets}, prediction logits~\cite{hinton2015distillingknowledgeneuralnetwork}, or activation boundaries~\cite{heo2019knowledge}]. This approach was originally put forth by Hinton et al.~\cite{hinton2015distilling} to supervise students based on the hard and soft label's output by the teacher, and nowadays there is a lot of work on using distillation for knowledge transfer to help the model get better performance. Earlier knowledge distillation (KD) techniques can be classified into three distinct categories: distillation from logits, distillation from features, and distillation based on attention. In terms of logit distillation, DIST~\cite{huang2022knowledge} employs the Pearson correlation coefficient in place of KL divergence, combining both inter- and intra-class relationships. SRRL~\cite{yang2021knowledge} ensures that the logits output from the teacher and the features of the student, after the teacher's linear layer, are identical. WSLD~\cite{zhou2021rethinking} examines soft labels and assigns varying weights to them based on the bias-variance trade-off. In addition to logit distillation, 
Several studies~\cite{cao2022pkd,shu2021channel,yang2022focal,yang2022vitkd} concentrate on transferring knowledge through intermediate features.
FitNet~\cite{romero2014fitnets} directly distills semantic information from these intermediate features. AT~\cite{zagoruyko2016paying} shifts the attention from feature maps to the student model. RKD~\cite{park2019relational} extracts relationships from the feature maps. MGD~\cite{yang2022masked} masks the features of the student model, compelling it to replicate the features of the teacher model. To our knowledge, AT~\cite{zagoruyko2016paying} is the sole knowledge distillation method that focuses on transferring attention, defining the attention map as a spatial representation that highlights the areas of the input that the model concentrates on the most. Wang et al. propose the HDETrack~\cite{wang2024HDETrack} which employs a hierarchical knowledge distillation strategy to augment the student tracking network from multi-modal or multi-view teacher network. 
In this paper, we employ both logits and intermediate features for knowledge distillation, believing that the integration of these two methods can greatly enhance knowledge transfer and improve the effectiveness of the student model.

\begin{figure*}
\centering
\includegraphics[width=0.8\linewidth]{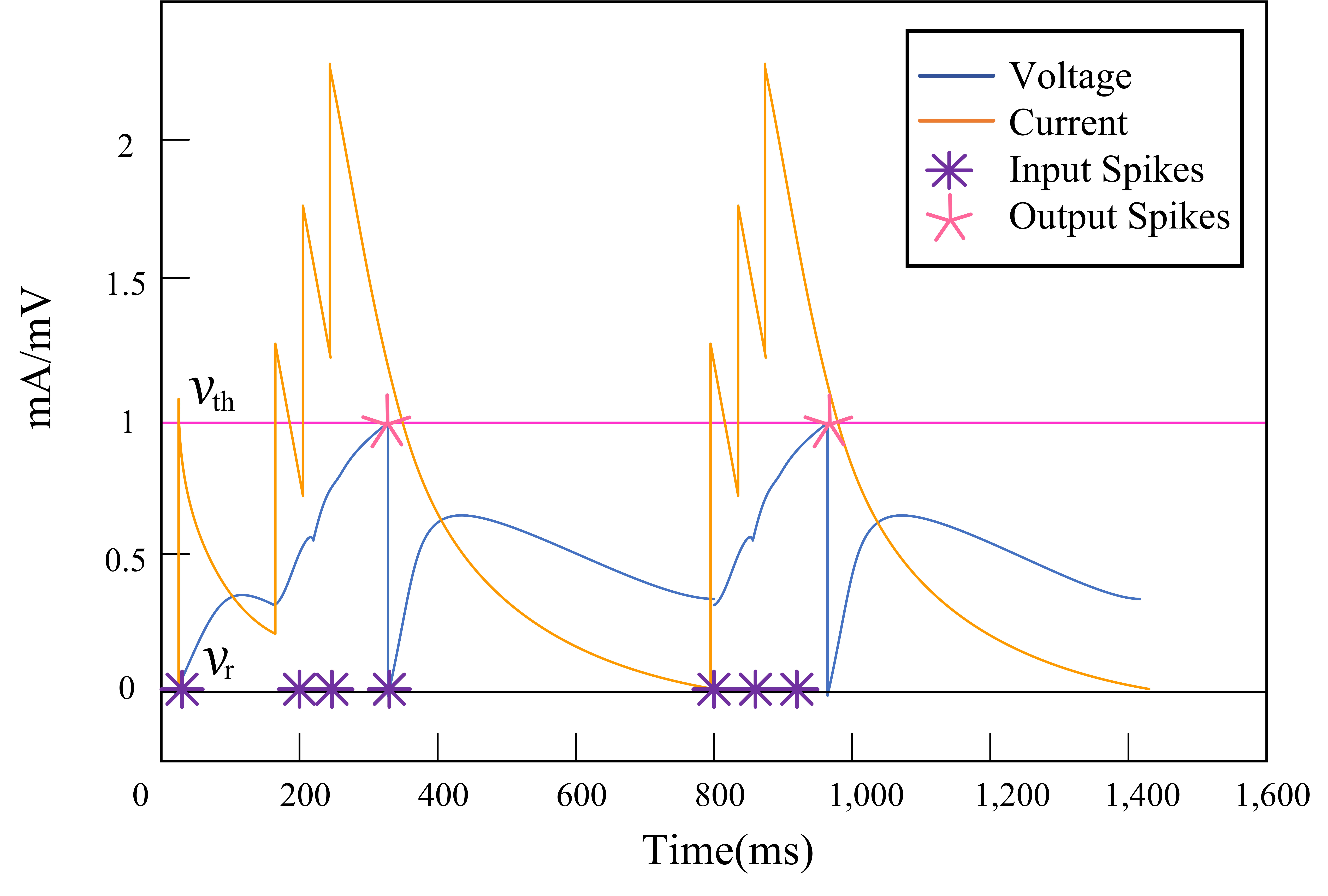}
\caption{
Simulation of the LIF model. Voltage and current rise with the onset of new spikes, resulting in the generation of an output spike if the voltage reaches $v_{th}$, after which it’s set to $v_{r}$, The diagram is re-drawn based on FPGA~\cite{moursi2024efficient}.} 
\label{LIF_model}
\end{figure*}

\section{Our Proposed Approach}

\subsection{Preliminaries: Leaky-Integrate-and-Fire (LIF) model}

The Leaky Integrate-and-Fire (LIF) model is a fundamental framework for simulating neuronal dynamics, effectively capturing the essential characteristics of biological neurons, as shown in Fig.~\ref{LIF_model}. The LIF model effectively simulates spiking neuron dynamics by integrating incoming signals while accounting for membrane leakage. Its simplicity and biological relevance establish it as a cornerstone in computational neuroscience, allowing researchers to investigate neuronal behavior and network dynamics. Additionally, its efficiency makes it suitable for applications in spiking neural networks, where energy consumption and computational resources are critical. This model comprises two key components: leaky integration and reset behavior. The main function of the Leaky Integration is represented by the following equation:
\begin{align}
\label{leaky_integration}
&\tau_m \frac{du}{dt} = -[u(t) - u_{rest}] + R I(t) 
\end{align}
In this equation,${u(t)}$ indicates the membrane potential of the neuron, urest refers to the resting membrane potential, $R$ represents the membrane resistance, and $I(t)$ signifies the input current. The term ${\tau_m}$ is the time constant, determining how quickly the membrane potential responds to inputs. When the neuron receives synaptic inputs, the membrane potential increases, integrating these signals over time. Nonetheless, because of the leaky characteristics of the membrane, the potential diminishes over time, indicating a gradual charge loss. Once the membrane potential hits a threshold $u_{th}$, the neuron generates an action potential, resulting in the reset of the membrane potential:
\begin{align}
\label{reset_behavior}
&u(t_f + \delta) = u_r, \quad \ u(t_f) \geq u_{th}.
\end{align}
Here, ${u(t_f)}$ is the membrane potential at the firing time, and $u_{r}$ is the reset potential. This reset behavior mimics the firing and refractory period of biological neurons, allowing the model to represent the spiking nature of neuronal activity accurately.

\begin{figure*}
\centering
\includegraphics[width=1\linewidth]{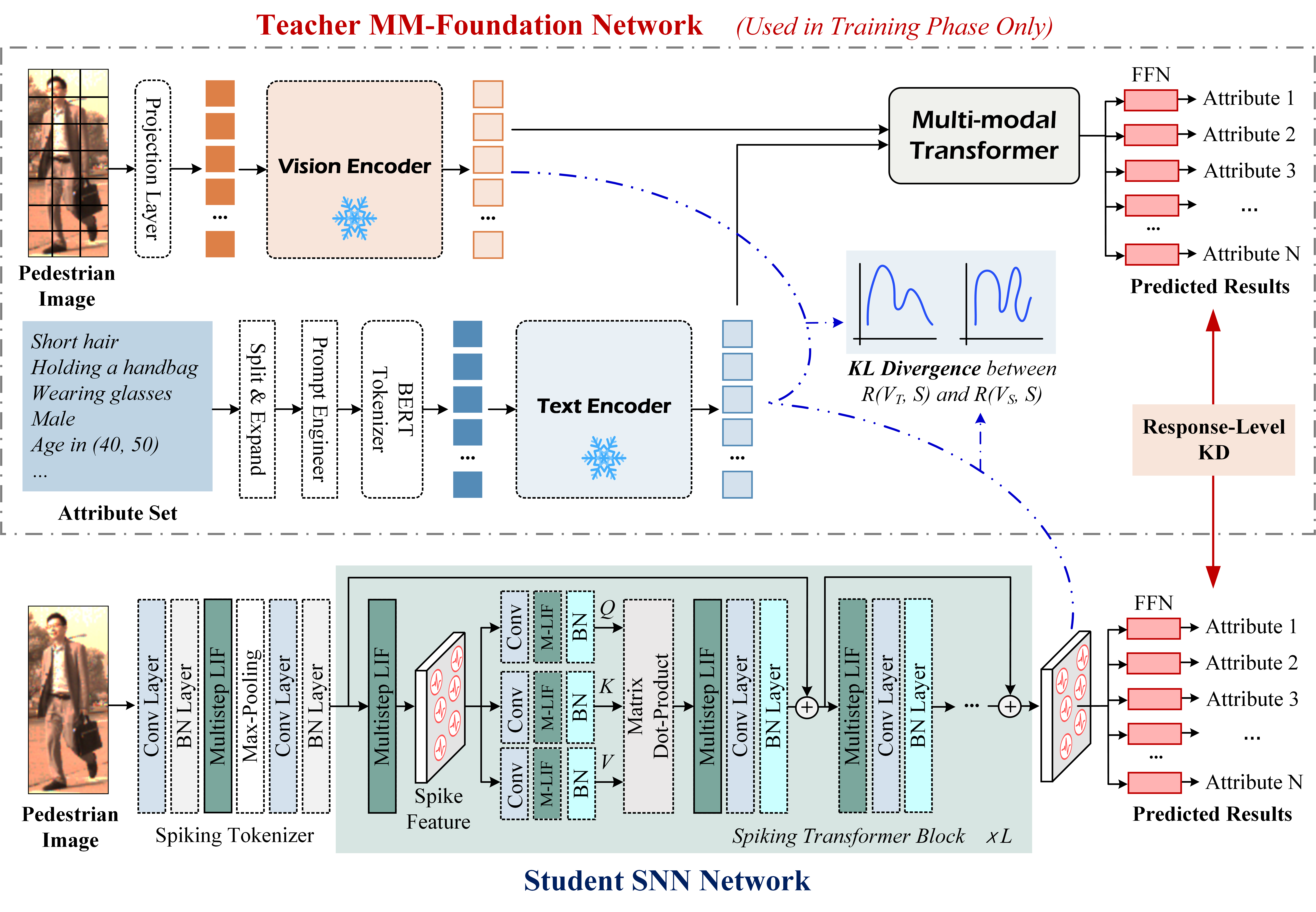}
\caption{ An overview of our proposed SNN-PAR framework, designed for energy-efficient pedestrian attribute recognition. 
}
\label{fig:framework}
\end{figure*}

\subsection{Overview} 
This paper presents a new deep learning framework called SNN-PAR, as shown in Fig.~\ref{fig:framework}. The core of this framework lies in harnessing the unique spatiotemporal dynamics and low-power characteristics of Spiking Neural Networks (SNNs) to achieve fine-grained image feature extraction while minimizing power consumption. SNNs simulate the information processing strategies of biological neural networks, enabling the efficient capture of key spatiotemporal features from images while maintaining energy efficiency. In addition, we use knowledge distillation techniques to improve the efficiency of feature learning and transferability. In this process, a pre-trained teacher model is employed to guide the learning of the SNN. The teacher model typically offers rich knowledge of image features. By transferring abstract high-level information from the teacher model to the SNN, our approach enhances the richness of the feature content while also bolstering their resilience. This is particularly important for dealing with complex and variable environmental conditions and noise interference.

\subsection{Network Architecture}

In this section, we will focus on the network architectures of our proposed SNN-PAR framework, including the Teacher and Student PAR module, Knowledge Distillation Enhanced Learning, and the loss functions used for the optimization. more details of these modules will be introduced in the subsequent paragraphs.

\noindent $\bullet$ \textbf{Teacher PAR Module.~} 
Current pedestrian attribute recognition models typically utilize Convolutional Neural Networks (CNNs) as the backbone, achieving remarkable performance. We adopt the VTB model~\cite{cheng2022VTB} as our teacher model, which excels in extracting image features, providing more comprehensive and precise knowledge to the student model. This enables the student model to improve its accuracy in attribute prediction.

Given a pedestrian image $\mathcal{X} \in \mathbb{R}^{C\times H \times W}$ and the corresponding annotated list of attributes $\mathcal{A} = \{ a_1, a_2, \ldots, a_M \}$, the image is first partitioned into multiple patches via a projection layer. These patches are subsequently encoded by the visual encoder from the teacher model, yielding visual features $\mathcal{F}^{t}_{v}$. Simultaneously, the attribute set $\mathcal{A}$ is expanded into sentence form and processed by the text encoder to generate textual features $\mathcal{F}_{t}$. VTB~\cite{cheng2022VTB} then introduces the Multi-modal Transform Fusion Module, which is specifically designed to effectively aggregate both text and visual features. This module employs advanced techniques to seamlessly integrate and enhance multimodal representations, enabling more comprehensive and precise feature fusion, which in turn improves model performance. The fused representation is then passed through a feed-forward network to produce the attribute prediction results $\mathcal{P}_{t}$. An overview of the Spikingformer pipeline is shown in Fig.~\ref{fig:framework}.

\noindent $\bullet$ \textbf{Student PAR Module.~} 
Spiking Neural Networks (SNNs) have gained significant attention in recent years owing to their biological relevance and exceptional energy efficiency, demonstrating robust computational capabilities in processing complex temporal information. Despite these advantages, the application of SNNs to pedestrian attribute recognition remains in early stages of exploration. To advance research in this area, we adopt the Spikingformer model~\cite{zhou2023spikingformer} as the student model to further investigate the specific use of SNNs for pedestrian attribute recognition. This work aims to provide deeper theoretical insights and practical guidance, contributing to the advancement and maturity of SNN applications in pedestrian attribute recognition.

The design of the student model, Spikingformer, comprises a Spiking Tokenizer (ST), a series of Spiking Transformer Blocks, and a Classification Head. input 2D image $\mathcal{I} \in \mathbb{R}^{C\times H \times W}$, here $C$, $H$, $W$ denotes the channel, height, and width of the image, respectively, the Spiking Tokenizer is applied for patch embedding. Specifically, the first layer serves as a spike encoder when processing the static images. As shown in Eq.~\ref{segmented image_1} and Eq.~\ref{segmented image_2}, the convolutional component of ConvBN refers to a 2D convolution layer, while MP and SN denote max pooling and multi-step spiking neurons, respectively. Spiking Patch Embedding (SPE) without downsampling is based on Eq.~\ref{segmented image_1}, and Spiking Patch Embedding with Downsampling (SPED) utilizes Eq.~\ref{segmented image_2}. 
Ultimately, the input $\mathcal{I}$ divided into a sequence of image patches $X \in \mathbb{R}^{N\times D}$, represented as:
\begin{align}
\label{segmented image_1} 
&{X}= ConvBN(SN(\mathcal{I})) \\
\label{segmented image_2} 
&{X} = ConvBN(MP(SN(\mathcal{I}))) 
\end{align} 
After passing through the Spiking Tokenizer, the spiking patches $X$ are processed by $L$ Spiking Transformer Blocks. In a manner similar to the standard ViT encoder block, each Spiking Transformer Block includes a Spiking Self-Attention (SSA) mechanism along with a Spiking MLP block. The SSA mechanism draws inspiration from the pure spiking self-attention design in Spikingformer. In Spikingformer~\cite{zhou2023spikingformer}, the spike-driven residual mechanism enhances SSA by repositioning the spiking neuron layer to avoid the multiplication of integer and floating-point weights, while replacing the LinearBN structure from Spikformer with ConvBN. Consequently, the Spiking Self-Attention (SSA) mechanism is defined as follows:
\begin{align}\label{1}
&X^\prime = SN(X), \\
&Q=SN_Q(ConvBN_Q(X^\prime)), \\
&K=SN_K(ConvBN_K(X^\prime)), \\
&V=SN_V(ConvBN_V(X^\prime)), \\
&SSA(Q, K, V) = ConvBN(SB(QK^TV*s)), 
\end{align}
where $Q$, $K$, and $V$ denote pure spike data, consisting exclusively of binary values ($0$, $1$). The scaling factor $s$, as described in~\cite{zhou2022spikformer}, modulates the magnitude of the matrix multiplication outputs. The Spiking MLP block integrates two Spiking Perceptron Ensembles (SPEs), as outlined in Eq.~\ref{segmented image_1}. The spiking Transformer block serves as a fundamental component of the Spikingformer architecture. We employ a fully connected layer as the classifier after the last Spiking Transformer module. The visual features $\mathcal{F}^{s}_{v}$ output from the Spiking Transformer Block is processed through the Classification Head to obtain the final prediction results $\mathcal{P}_{s}$. Finally, We adopt a weighted binary cross-entropy loss function to alleviate the distribution imbalance among pedestrian attributes, which is commonly employed in optimization for attribute recognition models. As a result, the formulation of our model's classification head is as follows:
\begin{equation}
    \mathcal{Y} = FC(AvgPooling({\mathcal{X}}_{L})). 
\end{equation}

\noindent $\bullet$ \textbf{Knowledge Distillation Enhanced Learning.~} In this study, we adopt a dual-level knowledge distillation strategy, conducting learning at both the feature and response levels. This hierarchical knowledge transfer framework ensures that the student model effectively assimilates knowledge from the teacher model, thereby enhancing its predictive performance. At the feature level, we focus on aligning the feature representations between the teacher visual features $\mathcal{F}^{t}_{v}$ and the student visual features $\mathcal{F}^{t}_{s}$, carefully tuning the spiking activity of the SNN to mimic the high-level features extracted by the teacher model in intermediate layers. At the response level, we focus on achieving the alignment of the predictions $\mathcal{P}_{t}$ of the teacher model and the predictions $\mathcal{P}_{s}$ of the student model. By transferring classification decision information from the teacher model to the student model, we ensure the accuracy of the SNN in its final predictions. This two-tier distillation approach not only enables the student model to inherit the teacher model’s strengths in feature extraction and decision-making but also improves the model’s generalization and robustness, all while maintaining low computational complexity. Additionally, this distillation method effectively mitigates issues such as gradient vanishing and overfitting, which are common in traditional SNN training, thus rendering the model more robust in processing real-world image data.

\subsection{Loss Function} 
In this study, we use a combination of cross-entropy loss $\mathcal{L}_{CE}$ and distillation loss functions $\mathcal{L}_{respKD}$, $\mathcal{L}_{featKD}$ to optimize the SNN-PAR framework. 
$\mathcal{L}_{CE}(S, y)$ is the cross-entropy loss between the output 
S of the student model and the labels y, which is typically defined as:
\begin{equation} 
\mathcal{L}_{CE}(S, y) = -\sum_{i} y_i \log(S_i). 
\end{equation}
$\mathcal{L}_{respKD}(S, T)$ is the response distillation loss between the output from the student model S and the output from the teacher model T, typically the Kullback-Leibler divergence, defined as: 
\begin{equation}
\mathcal{L}_{respKD}(S, T) = D_{KL}(T || S) = \sum_{i} T_i (\log(T_i) - \log(S_i)). 
\end{equation}
In this formula, \( T_i \) and \( S_i \) represent the \( i_{th} \) element of the output from the teacher model and the student model, typically obtained as probability distributions through the Softmax function. 
$\mathcal{L}_{featKD}$ is the feature distillation loss, defined as: 
\begin{equation}
\mathcal{L}_{featKD} = \sum_{i} P(\text{sim}(F_t, F_v))_i \log\left(\frac{P(\textbf{\(sim\)}(F_t, F_v'))_i}{P(\text{sim}(F_t, F_v))_i}\right). 
\end{equation}
Among them, \textbf{\( F_t \)} is the text feature of the teacher model, \textbf{ \( F_v \)} is the visual feature of the teacher model, and \textbf{\(F_v'\)} is the visual feature of the student model. 
\textit{sim} is the cosine similarity between two features.

Based on the above loss functions, the distillation loss is attached with weight coefficients $\alpha$ and $\beta$, respectively, along with the temperature coefficient T, which together form the final loss function:
\begin{equation}
\mathcal{L} = \mathcal{L}_{CE}+\alpha \mathcal{L}_{respKD}+\beta \mathcal{L}_{featKD}. 
\end{equation}

\section{Experiments} 

\subsection{Datasets and Evaluation Metric} 
To evaluate the effectiveness of our proposed SNN-PAR model, we perform experiments on three publicly accessible datasets: \textbf{PETA}~\cite{deng2014peta}, \textbf{PA100K}~\cite{2017pa100k} and \textbf{RAPv1}~\cite{2016rapv1}. 

\noindent $\bullet$  \textbf{PETA} comprises 19,000 images of pedestrians in outdoor or indoor settings, along with 61 binary attributes. These images are divided into training, validation, and testing subsets, containing 9,500, 1,900, and 7,600 images, respectively. In line with previous studies, we choose 35 pedestrian attributes for our experiments.

\noindent $\bullet$  \textbf{PA100K} is the most extensive for pedestrian attribute recognition, encompassing 100,000 pedestrian images with 26 binary attributes. Note that, 90,000 of these images are designated for training and validation purposes, while a separate set of 10,000 images is reserved for testing.

\noindent $\bullet$  \textbf{RAPv1} comprises 41,585 pedestrian images and 69 binary attributes, 33,268 images are designated for training. Typically, 51 attributes are selected for both training and evaluation purposes.

In our experiments, we utilize five widely recognized evaluation metrics to measure performance: 
\textbf{mean Accuracy}~(mA), 
\textbf{Accuracy}~(Acc), 
\textbf{Precision}~(Prec), 
\textbf{Recall} and 
\textbf{F1-score}~(F1),
these metrics are defined as follows:
\begin{equation}
     Accuracy=\dfrac{TP+TN}{FP+FN+TP+TN}
 \end{equation}
 \begin{equation}
     Precision=\dfrac{TP}{FP+TP}, ~~~~ Recall=\dfrac{TP}{TP+FN}
 \end{equation}
\begin{equation}
    F1-score=\dfrac{2\times Recall\times Precision }{Recall+Precision} 
\end{equation}
Where $TP$ denotes the number of samples that were correctly predicted as positive (true positives), $TN$  represents the count of samples accurately identified as negative (true negatives). Additionally, $FP$ refers to the number of false positives, which are samples incorrectly predicted as positive, while $FN$ indicates the number of false negatives, or samples that were incorrectly classified as negative.


\subsection{Implementation Details} 

In the training stage, we utilize a batch size of 12 and proceed to train the model over a complete duration of 60 epochs. In the teacher model, the input to the visual encoder of the student model and teacher model is set to $256 \times 128$.  This configuration is consistently applied across experiments on the RAPv1, PETA, and PA100K datasets. The initial learning rate is set at 8e-4, with a decay rate of 1e-4 as training progresses. We use the Adam optimizer~\cite{diederik2014adam} for our experiments. To optimize the learning process, we implement a warm-up strategy, gradually increasing the learning rate from 0 to an initial value of 1e-3 over the first 10 epochs. As the iteration count mounts, we decrease the learning rate by a multiplicative factor of 0.1. Knowledge distillation is performed with a temperature coefficient set to 2. Further details are available in our source code.


\subsection{Comparison on Public Benchmarks}

\begin{table*}[!htb]
\center
\caption{Comparison with SOTA methods on PETA, PA100K and RAPv1 datasets.} \label{Comparisononpublicdatasets} 
\resizebox{1\textwidth}{!}{
\begin{tabular}{l|ccccc|ccccc|ccccc}
\hline \toprule [0.5 pt] 
\multicolumn{1}{c|}{\multirow{2}{*}{Methods}} & 
\multicolumn{5}{c|}{PETA} 
& \multicolumn{5}{c|}{PA100K} 
& \multicolumn{5}{c}{RAPv1} \\ 
\cline{2-16} 
  \multicolumn{1}{c|}{} &
  \multicolumn{1}{c}{mA} &
  \multicolumn{1}{c}{Acc} & 
  \multicolumn{1}{c}{Prec} &
  \multicolumn{1}{c}{Recall} &
  \multicolumn{1}{c|}{F1} &
  \multicolumn{1}{c}{mA} &
  \multicolumn{1}{c}{Acc} &
  \multicolumn{1}{c}{Prec} &
  \multicolumn{1}{c}{Recall} &
  \multicolumn{1}{c|}{F1}  &
  \multicolumn{1}{c}{mA} &
  \multicolumn{1}{c}{Acc} &
  \multicolumn{1}{c}{Prec} &
  \multicolumn{1}{c}{Recall} &
  \multicolumn{1}{c}{F1}  \\ \hline
SSCsoft~\cite{2021ssc} & 86.52 & 78.95 & 86.02 & 87.12 & 86.99 & 81.87 & 78.89 & 85.98 & 89.10 & 86.87 & 82.77 & 68.37 & 75.05 & 87.49 & 80.43\\ 
IAA~\cite{2022iaacaps}  & 85.27 & 78.04 & 86.08 & 85.80 & 85.64 & 81.94 & 80.31 & 88.36 & 88.01 & 87.80 & 81.72 & 68.47 & 79.56 & 82.06 & 80.37\\
MCFL~\cite{Chen2022MCFL}  & 86.83 & 78.89 & 84.57 & 88.84 & 86.65 & 81.53 & 77.80 & 85.11 & 88.20 & 86.62 & 84.04 & 67.28 & 73.44 &87.75 & 79.96\\
DRFormer~\cite{2022drformer} & 89.96 &81.30 & 85.68 &91.08 &88.30 & 82.47 & 80.27 & 87.60 & 88.49 & 88.04 & 81.81 &70.60 & 80.12 & 82.77 & 81.42\\
VAC~\cite{guo2022visual}   & - & - & - & - & - & 82.19 & 80.66 & 88.72 & 88.10 & 88.41 & 81.30 & 70.12 &81.56 & 81.51 &81.54\\
DAFL~\cite{jia2022learning}  & 87.07 & 78.88 & 85.78 & 87.03 & 86.40 & 83.54 & 80.13 & 87.01 & 89.19 & 88.09 & 83.72 & 68.18 & 77.41 & 83.39 & 80.29  \\
CGCN~\cite{Fan2022CGCN}  & 87.08 & 79.30 & 83.97 & 89.38 & 86.59 & - & - & - & - & - & 84.70 & 54.40 & 60.03 & 83.68 & 70.49 \\ 
CAS~\cite{yang2021cascaded} & 86.40 & 79.93 & 87.03 & 87.33 & 87.18 & 82.86 & 79.64 & 86.81 & 87.79 & 85.18 & 84.18 & 68.59 & 77.56 & 83.81 & 80.56\\  
VTB~\cite{cheng2022VTB}  & 85.31 & 79.60 & 86.76 & 87.17 & 86.71 & 83.72 & 80.89 & 87.88 & 89.30 & 88.21 & 82.67 & 69.44 & 78.28 & 84.39 & 80.84 \\
\hline
SNN-PAR (Ours) & 80.58 & 73.55 & 81.76 & 82.79&  81.96 &  73.86 & 71.70 & 83.03 &  81.30 &  81.67  &  75.43 &  63.06& 74.67 &  78.28 &  75.94 \\
\hline \toprule [0.5 pt] 
\end{tabular}}
\end{table*}

We evaluate the proposed SNN-PAR model against state-of-the-art methods on three benchmark datasets: PA100K~\cite{2017pa100k}, RAPv1~\cite{2016rapv1}, and PETA~\cite{deng2014peta}. These datasets, renowned for their diverse pedestrian attributes and challenging scenarios, provide a robust foundation for assessing the effectiveness of attribute recognition models under real-world conditions.




The results on three public datasets are provided in Table~\ref{Comparisononpublicdatasets}. For the PA100K~\cite{2017pa100k} dataset, our proposed SNN-PAR model achieves 73.86 in mA, 71.70 in Accuracy, 83.03 in Precision, 81.30 in Recall, and 81.67 in F1-score. On the RAPv1~\cite{2016rapv1} dataset, it records 75.43 for mA, 63.06 for Accuracy, 74.67 for Precision, 78.28 for Recall, and 75.94 for F1-score. Lastly, the evaluation on the PETA~\cite{deng2014peta} dataset results in 80.58, 73.55, 81.76, 82.79, and 81.96 for mA, Accuracy, Precision, Recall, and F1-score, respectively.

As shown in Table~\ref{Comparisononpublicdatasets}, while most prior methods outperform our SNN-PAR framework by approximately 4 to 5 points, this is expected given the superior performance of ANN architectures. In contrast, our model prioritizes balancing accuracy with energy efficiency, achieving comparable performance while consuming significantly less energy.

\subsection{Ablation Studies} 

\noindent $\bullet$ \textbf{Baseline Comparison: SNN vs. Transformer-based PAR Model.~}
To assess the efficiency and effectiveness of the proposed SNN-PAR model, we perform a comparative analysis by substituting the SNN module in the student model with a Transformer (ViT) module. This comparison is designed to highlight the benefits of employing Spiking Neural Networks (SNNs) over conventional Transformer architectures, focusing on both energy efficiency and attribute recognition accuracy.
\textbf{SNN-PAR:} Our proposed model incorporates the SNN module in the student branch.
\textbf{Transformer-based model:} A variant of the SNN-PAR model, where the SNN module in the student branch is replaced with a Transformer (ViT) module. Both models are trained on the PA100K pedestrian attribute recognition dataset to ensure a fair and consistent evaluation. Table~\ref{comparative_analysis} presents the performance comparison in terms of Acc, mA, Prec, Rec, and F1.

\FloatBarrier
\begin{table*}[!htbp]
\centering
\small   
\caption{Comparative analysis on  PA100K dataset}  
\label{comparative_analysis}
\begin{tabular}{c|c|c|c|c|c}
\hline \toprule [0.5 pt] 
Backbone & mA& Acc & Prec & Recall &F1 \\
\hline
ViT &80.33 &78.24 &86.48 &87.48 &86.49 \\
\hline
SNN &73.86  &71.70  &83.03 	&81.30		&81.67 \\
\hline \toprule [0.5 pt] 
\end{tabular}
\end{table*}
\FloatBarrier

As presented in Table~\ref{comparative_analysis}, the SNN-PAR model achieves scores of 73.86, 71.70, 83.03, 81.30, and 81.67 for mA, Accuracy, Precision, Recall, and F1, respectively. In comparison, the Transformer-based model attains higher values of 80.33, 78.24, 86.48, 87.48, and 86.49 across the same metrics. While it is expected that the ViT model exhibits superior performance due to its more complex architecture and higher energy consumption, our SNN-PAR model strikes a balance by sacrificing a small degree of accuracy in favor of greater energy efficiency and a more lightweight network architecture. 

\begin{table*} 
\center
\small 
\caption{SNN with knowledge distillation on the PA100K dataset} 
\label{distillation} 
\begin{tabular}{c|ccc|ccc}
\hline \toprule [0.5 pt]
\multicolumn{1}{c|}{NO.} & 
\multicolumn{1}{c}{SNN}  &
\multicolumn{1}{c}{$\mathcal{L}_{featKD}$}  &
\multicolumn{1}{c|}{$\mathcal{L}_{respKD}$} & 
\multicolumn{1}{c}{mA} & 
\multicolumn{1}{c}{F1} \\ 
\hline 
1   &\checkmark & &  & {73.86}  &{81.67}    \\ 
2   &\checkmark & \checkmark & &75.10   &82.34    \\
3   &\checkmark & & \checkmark  &74.27  &82.52    \\
4   &\checkmark & \checkmark  & \checkmark  &74.76 &83.16  \\
\hline \toprule [0.5 pt] 
 \end{tabular}
\end{table*}

\noindent $\bullet$ \textbf{Effects of Knowledge Distillation.~} 
At this stage, our objective is to refine the student model’s acquisition of knowledge from the teacher model by employing knowledge distillation techniques, incorporating both feature-level and response-level distillation. We conduct experiments using each distillation method independently to assess their effectiveness. The results of these experiments are presented in the following section.

\textbf{SNN with feature-level distillation:} We also conduct an experiment to evaluate the effectiveness of feature-level distillation (second row). As illustrated in Table~\ref{distillation}, compared to the original SNN model (first row), the SNN model with the additional feature-level distillation strategy achieves improvements of +1.24 in mA and +0.67 in Accuracy.

\textbf{SNN with response-level distillation:} As shown in Table~\ref{distillation}, we initially apply only response-level distillation to validate its effectiveness (third row). Notably, the  mA and F1 scores improve by +0.41,+0.85, respectively, compared to the baseline SNN model (first row), highlighting the importance of the proposed response-level distillation.

\textbf{SNN with response and feature Distillation:} To further improve the performance of our student model, we combine the aforementioned distillation strategies. As shown in Table~\ref{distillation}, the SNN model with both response-level and feature-level distillation (fourth row) achieves the highest performance, with scores of 74.76 in mA and 83.16 in F1. This demonstrates the effectiveness of integrating these two levels of distillation.









\begin{table}
\center
\small  
\caption{Comparison on different model.} 
\label{parameter_analysis} 
\resizebox{0.5\linewidth}{!}{ 
\begin{tabular}{c|c|c}
\hline \toprule [0.5 pt]
\multicolumn{1}{c|}{Method}
&\multicolumn{1}{c|}{Backbone}
&\multicolumn{1}{c}{Parameter(M)}\\ \hline
DFDT\cite{ZHENG2023105708} & Swin-B   & 87.59  \\  
VTB\cite{cheng2022VTB} & ViT-B/16      & 157.54 \\
PromtPAR\cite{wang2024pedestrian} & ViT-L/14     & 435.93  \\
\hline
Ours & SNN     & 65.59   \\
\hline \toprule [0.5 pt] 
\end{tabular}} 
\end{table}

\subsection{Parameter Analysis}
In this section, we present the key parameters of our SNN. Specifically, we report the total number of learnable parameters in the model. As illustrated in Table~\ref{parameter_analysis}, Comparison with 157.54M learnable parameters for ViT-B/16 and 87.59M learnable parameters for Swin-B, our SNN-PAR model contains only 65.59M learnable parameters, making the overall network significantly more lightweight. In particular, the number of parameters in our method is more than halved compared to the ViT-B/16 method.

\subsection{Visualization} 
In this section, we present a case study highlighting successfully predicted attributes on the PA100K dataset. To offer a clearer insight into the prediction process of our model, we also include heatmap visualizations that illustrate the predicted regions of interest.


\begin{figure*} 
\centering
\small
\includegraphics[width=1\linewidth]{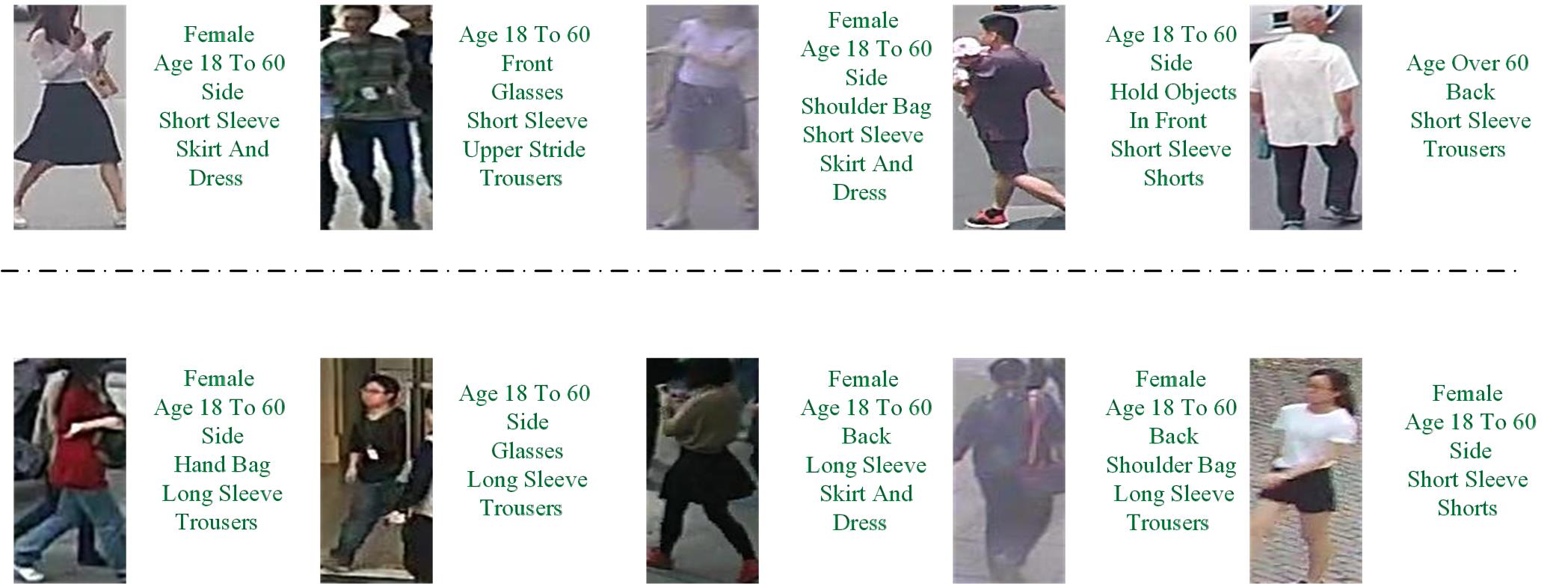}
\caption{Visualization of pedestrian attributes predicted by our proposed model. The \textcolor{SeaGreen4}{\emph{green}} attributes are corrected predicted ones.} 
\label{right_predicted} 
\end{figure*}

\noindent $\bullet$ \textbf{Attributes Predicted using Our Student Model.}  
As depicted in Fig.~\ref{right_predicted}, we showcase 10 predictions generated by our model on the PA100K dataset. It is clear that the model effectively identifies a range of pedestrian attributes, including gender, age, motion, and outfit, among others.

\noindent $\bullet$ \textbf{Heatmap Visualization.~} 
To deliver a more straightforward visualization of the key areas attended to by our model on the PA100K dataset when predicting pedestrian attributes, we visualize the model’s prediction process using heatmaps. As shown in Fig.~\ref{heatmap}, the model accurately focuses on the relevant regions corresponding to the pedestrian attributes during prediction.

\subsection{Limitation Analysis} 
\begin{figure*}
\centering
\small
\includegraphics[width=1\linewidth]{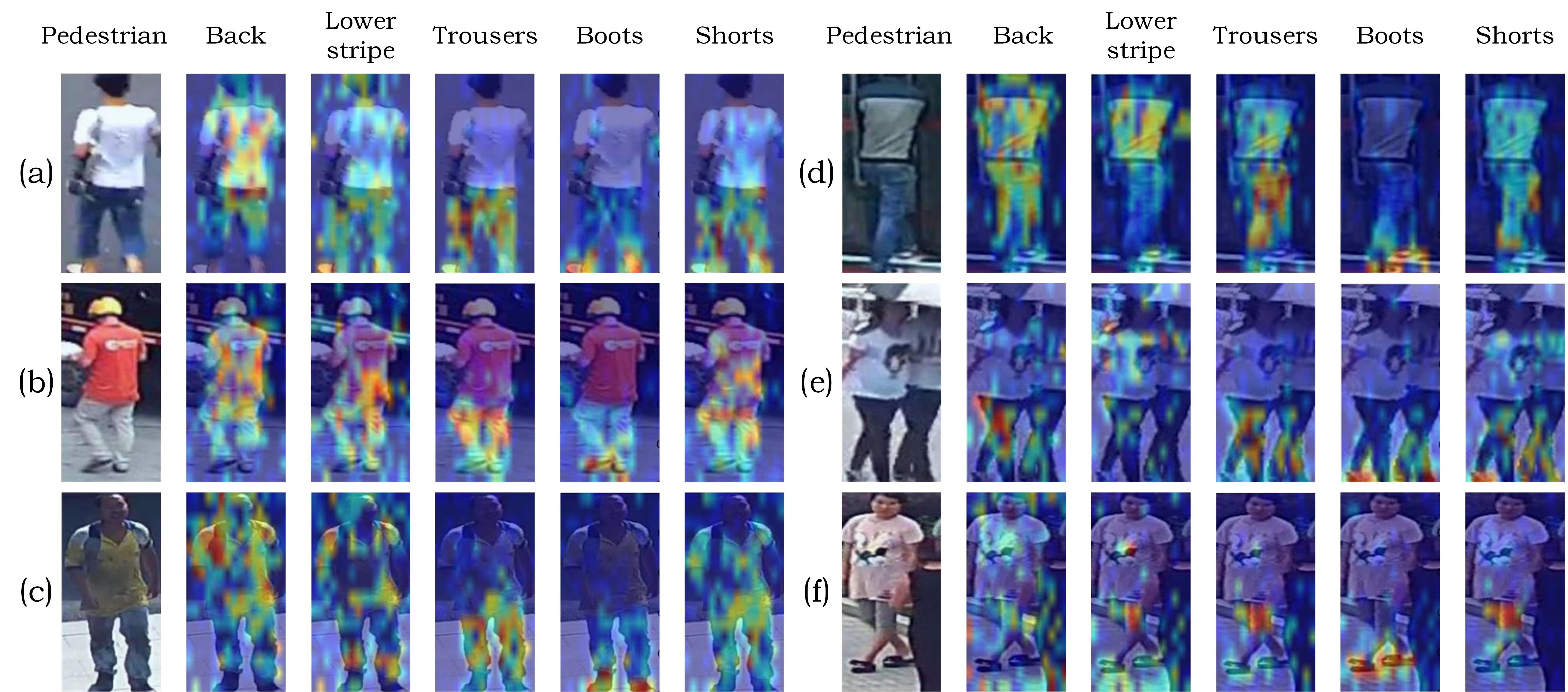}
\caption{Visualization of heat maps given the corresponding pedestrian attribute.} 
\label{heatmap} 
\end{figure*}

As shown in Fig.~\ref{heatmap}, while our model focuses on broad regions within pedestrian images, such as those corresponding to motion and outfit, the localization is not sufficiently precise. Moreover, as reflected in Table~\ref{Comparisononpublicdatasets}, the performance of our SNN-PAR model, while more energy-efficient and lightweight, falls short in accuracy compared to other state-of-the-art methods. In our future work, we plan to explore the design of a hybrid ANN-SNN architecture to strike a better balance between accuracy and energy efficiency.

\section{Conclusion} 
In this paper, we propose a novel SNN-based pedestrian attribute recognition framework, termed SNN-PAR, leveraging Spiking Neural Networks (SNNs) to achieve more energy-efficient performance. However, general SNN-based models often struggle to deliver high accuracy. To strike a balance between accuracy and energy efficiency, we employ a teacher-student model to train our SNN student model. We incorporate two levels of distillation—response-level and feature-level, which significantly enhance the attribute recognition accuracy of our SNN-PAR model. While our framework performs well on three benchmark datasets, there remains a notable gap compared to methods using purely ANN architectures. In future research, we intend to investigate more efficient frameworks to enhance the performance of our model further.

\bibliographystyle{plain}

\bibliography{reference}

\begin{thebibliography}{10}

\bibitem{abdulnabi2015multi}
Abrar~H Abdulnabi, Gang Wang, Jiwen Lu, and Kui Jia.
\newblock Multi-task cnn model for attribute prediction.
\newblock {\em IEEE Transactions on Multimedia}, 17(11):1949--1959, 2015.

\bibitem{Chen2022MCFL}
Lin~Chen andJingkuan Song andXuerui Zhang~andMingsheng Shang.
\newblock Mcfl: multi-label contrastive focal loss for deep imbalanced
  pedestrian attribute recognition.
\newblock {\em Neural Computing and Applications}, 2022.

\bibitem{bing2019end}
Zhenshan Bing, Zhuangyi Jiang, Long Cheng, Caixia Cai, Kai Huang, and Alois
  Knoll.
\newblock End to end learning of a multi-layered snn based on r-stdp for a
  target tracking snake-like robot.
\newblock In {\em 2019 International Conference on Robotics and Automation
  (ICRA)}, pages 9645--9651. IEEE, 2019.

\bibitem{burelo2020spikingneuralnetworksnn}
Karla Burelo, Mohammadali Sharifshazileh, Niklaus Krayenbühl, Georgia
  Ramantani, Giacomo Indiveri, and Johannes Sarnthein.
\newblock A spiking neural network (snn) for detecting high frequency
  oscillations (hfos) in the intraoperative ecog, 2020.

\bibitem{cao2022pkd}
Weihan Cao, Yifan Zhang, Jianfei Gao, Anda Cheng, Ke~Cheng, and Jian Cheng.
\newblock Pkd: General distillation framework for object detectors via pearson
  correlation coefficient.
\newblock {\em Advances in Neural Information Processing Systems},
  35:15394--15406, 2022.

\bibitem{cao2015spiking}
Yongqiang Cao, Yang Chen, and Deepak Khosla.
\newblock Spiking deep convolutional neural networks for energy-efficient
  object recognition.
\newblock {\em International Journal of Computer Vision}, 113:54--66, 2015.

\bibitem{caporale2008spike}
Natalia Caporale and Yang Dan.
\newblock Spike timing--dependent plasticity: a hebbian learning rule.
\newblock {\em Annu. Rev. Neurosci.}, 31(1):25--46, 2008.

\bibitem{cheng2022VTB}
Xinhua Cheng, Mengxi Jia, Qian Wang, and Jian Zhang.
\newblock A simple visual-textual baseline for pedestrian attribute
  recognition.
\newblock {\em IEEE Transactions on Circuits and Systems for Video Technology},
  32(10):6994--7004, 2022.

\bibitem{chung2014gru}
Junyoung Chung, Caglar Gulcehre, KyungHyun Cho, and Yoshua Bengio.
\newblock Empirical evaluation of gated recurrent neural networks on sequence
  modeling.
\newblock {\em arXiv preprint arXiv:1412.3555}, 2014.

\bibitem{deng2014peta}
Yubin Deng, Ping Luo, Chen~Change Loy, and Xiaoou Tang.
\newblock Pedestrian attribute recognition at far distance.
\newblock In {\em Proceedings of the 22nd ACM international conference on
  Multimedia}, pages 789--792, 2014.

\bibitem{diederik2014adam}
P~Kingma Diederik.
\newblock Adam: A method for stochastic optimization.
\newblock {\em (No Title)}, 2014.

\bibitem{DosovitskiyViT}
Alexey Dosovitskiy, Lucas Beyer, Alexander Kolesnikov, Dirk Weissenborn,
  Xiaohua Zhai, Thomas Unterthiner, Mostafa Dehghani, Matthias Minderer, Georg
  Heigold, Sylvain Gelly, Jakob Uszkoreit, and Neil Houlsby.
\newblock An image is worth 16x16 words: Transformers for image recognition at
  scale.
\newblock In {\em International Conference on Learning Representations}, 2021.

\bibitem{duwek2021image}
Hadar~Cohen Duwek, Albert Shalumov, and Elishai~Ezra Tsur.
\newblock Image reconstruction from neuromorphic event cameras using
  laplacian-prediction and poisson integration with spiking and artificial
  neural networks.
\newblock In {\em Proceedings of the IEEE/CVF Conference on Computer Vision and
  Pattern Recognition}, pages 1333--1341, 2021.

\bibitem{eshraghian2023trainingspikingneuralnetworks}
Jason~K. Eshraghian, Max Ward, Emre Neftci, Xinxin Wang, Gregor Lenz, Girish
  Dwivedi, Mohammed Bennamoun, Doo~Seok Jeong, and Wei~D. Lu.
\newblock Training spiking neural networks using lessons from deep learning,
  2023.

\bibitem{Fan2022CGCN}
Haonan Fan, Hai-Miao Hu, Shuailing Liu, Weiqing Lu, and Shiliang Pu.
\newblock Correlation graph convolutional network for pedestrian attribute
  recognition.
\newblock {\em IEEE Transactions on Multimedia}, 24:49--60, 2022.

\bibitem{fan2023parformer}
Xinwen Fan, Yukang Zhang, Yang Lu, and Hanzi Wang.
\newblock Parformer: transformer-based multi-task network for pedestrian
  attribute recognition.
\newblock {\em IEEE Transactions on Circuits and Systems for Video Technology},
  2023.

\bibitem{fang2021incorporating}
Wei Fang, Zhaofei Yu, Yanqi Chen, Timoth{\'e}e Masquelier, Tiejun Huang, and
  Yonghong Tian.
\newblock Incorporating learnable membrane time constant to enhance learning of
  spiking neural networks.
\newblock In {\em Proceedings of the IEEE/CVF international conference on
  computer vision}, pages 2661--2671, 2021.

\bibitem{guo2022visual}
Hao Guo, Xiaochuan Fan, and Song Wang.
\newblock Visual attention consistency for human attribute recognition.
\newblock {\em International Journal of Computer Vision}, 130(4):1088--1106,
  2022.

\bibitem{he2016resnet}
Kaiming He, Xiangyu Zhang, Shaoqing Ren, and Jian Sun.
\newblock Deep residual learning for image recognition.
\newblock In {\em Proceedings of the IEEE conference on computer vision and
  pattern recognition}, pages 770--778, 2016.

\bibitem{heo2019knowledge}
Byeongho Heo, Minsik Lee, Sangdoo Yun, and Jin~Young Choi.
\newblock Knowledge transfer via distillation of activation boundaries formed
  by hidden neurons.
\newblock In {\em Proceedings of the AAAI conference on artificial
  intelligence}, volume~33, pages 3779--3787, 2019.

\bibitem{hinton2015distilling}
Geoffrey Hinton.
\newblock Distilling the knowledge in a neural network.
\newblock {\em arXiv preprint arXiv:1503.02531}, 2015.

\bibitem{hinton2015distillingknowledgeneuralnetwork}
Geoffrey Hinton, Oriol Vinyals, and Jeff Dean.
\newblock Distilling the knowledge in a neural network, 2015.

\bibitem{hochreiter1997lstm}
Sepp Hochreiter and J{\"u}rgen Schmidhuber.
\newblock Long short-term memory.
\newblock {\em Neural computation}, 9(8):1735--1780, 1997.

\bibitem{huang2022knowledge}
Tao Huang, Shan You, Fei Wang, Chen Qian, and Chang Xu.
\newblock Knowledge distillation from a stronger teacher.
\newblock {\em Advances in Neural Information Processing Systems},
  35:33716--33727, 2022.

\bibitem{huang2024attPersonRetrieval}
Yan Huang, Zhang Zhang, Qiang Wu, Yi~Zhong, and Liang Wang.
\newblock Attribute-guided pedestrian retrieval: Bridging person re-id with
  internal attribute variability.
\newblock In {\em Proceedings of the IEEE/CVF Conference on Computer Vision and
  Pattern Recognition}, pages 17689--17699, 2024.

\bibitem{2021ssc}
Jian {Jia}, Xiaotang {Chen}, and Kaiqi {Huang}.
\newblock {Spatial and Semantic Consistency Regularizations for Pedestrian
  Attribute Recognition}.
\newblock {\em arXiv e-prints}, page arXiv:2109.05686, September 2021.

\bibitem{jia2022learning}
Jian Jia, Naiyu Gao, Fei He, Xiaotang Chen, and Kaiqi Huang.
\newblock Learning disentangled attribute representations for robust pedestrian
  attribute recognition.
\newblock {\em Proceedings of the AAAI Conference on Artificial Intelligence},
  36(1):1069--1077, Jun. 2022.

\bibitem{kabilan2021neuromorphic}
R~Kabilan and N~Muthukumaran.
\newblock A neuromorphic model for image recognition using snn.
\newblock In {\em 2021 6th International Conference on Inventive Computation
  Technologies (ICICT)}, pages 720--725. IEEE, 2021.

\bibitem{deepmar}
Dangwei Li, Xiaotang Chen, and Kaiqi Huang.
\newblock Multi-attribute learning for pedestrian attribute recognition in
  surveillance scenarios.
\newblock In {\em 2015 3rd IAPR Asian Conference on Pattern Recognition
  (ACPR)}, pages 111--115, 2015.

\bibitem{2016rapv1}
Dangwei {Li}, Zhang {Zhang}, Xiaotang {Chen}, Haibin {Ling}, and Kaiqi {Huang}.
\newblock {A Richly Annotated Dataset for Pedestrian Attribute Recognition}.
\newblock {\em arXiv e-prints}, page arXiv:1603.07054, March 2016.

\bibitem{li2019visual}
Qiaozhe Li, Xin Zhao, Ran He, and Kaiqi Huang.
\newblock Visual-semantic graph reasoning for pedestrian attribute recognition.
\newblock In {\em Proceedings of the AAAI conference on artificial
  intelligence}, volume~33, pages 8634--8641, 2019.

\bibitem{li2024attmot}
Yunhao Li, Zhen Xiao, Lin Yang, Dan Meng, Xin Zhou, Heng Fan, and Libo Zhang.
\newblock Attmot: improving multiple-object tracking by introducing auxiliary
  pedestrian attributes.
\newblock {\em IEEE transactions on neural networks and learning systems},
  2024.

\bibitem{lin2019improving}
Yutian Lin, Liang Zheng, Zhedong Zheng, Yu~Wu, Zhilan Hu, Chenggang Yan, and
  Yi~Yang.
\newblock Improving person re-identification by attribute and identity
  learning.
\newblock {\em Pattern recognition}, 95:151--161, 2019.

\bibitem{2017pa100k}
Xihui {Liu}, Haiyu {Zhao}, Maoqing {Tian}, Lu~{Sheng}, Jing {Shao}, Shuai {Yi},
  Junjie {Yan}, and Xiaogang {Wang}.
\newblock {HydraPlus-Net: Attentive Deep Features for Pedestrian Analysis}.
\newblock {\em arXiv e-prints}, page arXiv:1709.09930, September 2017.

\bibitem{lu2023oagcn}
Wei-Qing Lu, Hai-Miao Hu, Jinzuo Yu, Yibo Zhou, Hanzi Wang, and Bo~Li.
\newblock Orientation-aware pedestrian attribute recognition based on graph
  convolution network.
\newblock {\em IEEE Transactions on Multimedia}, 26:28--40, 2024.

\bibitem{moursi2024efficient}
Mohamed Moursi, Jonas Ney, Bilal Hammoud, and Norbert Wehn.
\newblock Efficient fpga implementation of an optimized snn-based dfe for
  optical communications.
\newblock {\em arXiv preprint arXiv:2409.08698}, 2024.

\bibitem{national1931proceedings}
India National Academy~of Sciences.
\newblock Proceedings of the national academy of sciences.
\newblock National Acad. of Sciences, 1931.

\bibitem{paredes2019unsupervised}
Federico Paredes-Vall{\'e}s, Kirk~YW Scheper, and Guido~CHE De~Croon.
\newblock Unsupervised learning of a hierarchical spiking neural network for
  optical flow estimation: From events to global motion perception.
\newblock {\em IEEE transactions on pattern analysis and machine intelligence},
  42(8):2051--2064, 2019.

\bibitem{park2017attribute}
Seyoung Park, Bruce~Xiaohan Nie, and Song-Chun Zhu.
\newblock Attribute and-or grammar for joint parsing of human pose, parts and
  attributes.
\newblock {\em IEEE transactions on pattern analysis and machine intelligence},
  40(7):1555--1569, 2017.

\bibitem{park2019relational}
Wonpyo Park, Dongju Kim, Yan Lu, and Minsu Cho.
\newblock Relational knowledge distillation.
\newblock In {\em Proceedings of the IEEE/CVF conference on computer vision and
  pattern recognition}, pages 3967--3976, 2019.

\bibitem{romero2014fitnets}
Adriana Romero, Nicolas Ballas, Samira~Ebrahimi Kahou, Antoine Chassang, Carlo
  Gatta, and Yoshua Bengio.
\newblock Fitnets: Hints for thin deep nets.
\newblock {\em arXiv preprint arXiv:1412.6550}, 2014.

\bibitem{shen2022backpropagation}
Guobin Shen, Dongcheng Zhao, and Yi~Zeng.
\newblock Backpropagation with biologically plausible spatiotemporal adjustment
  for training deep spiking neural networks.
\newblock {\em Patterns}, 3(6), 2022.

\bibitem{shu2021channel}
Changyong Shu, Yifan Liu, Jianfei Gao, Zheng Yan, and Chunhua Shen.
\newblock Channel-wise knowledge distillation for dense prediction.
\newblock In {\em Proceedings of the IEEE/CVF International Conference on
  Computer Vision}, pages 5311--5320, 2021.

\bibitem{sung2020training}
Mingyu Sung and Yongtae Kim.
\newblock Training spiking neural networks with an adaptive leaky
  integrate-and-fire neuron.
\newblock In {\em 2020 IEEE international conference on consumer
  electronics-Asia (ICCE-Asia)}, pages 1--2. IEEE, 2020.

\bibitem{tang2022drformer}
Zengming Tang and Jun Huang.
\newblock Drformer: Learning dual relations using transformer for pedestrian
  attribute recognition.
\newblock {\em Neurocomputing}, 497:159--169, 2022.

\bibitem{2022drformer}
Zengming Tang and Jun Huang.
\newblock Drformer: Learning dual relations using transformer for pedestrian
  attribute recognition.
\newblock {\em Neurocomputing}, 497:159--169, 2022.

\bibitem{tian2015pedestrian}
Yonglong Tian, Ping Luo, Xiaogang Wang, and Xiaoou Tang.
\newblock Pedestrian detection aided by deep learning semantic tasks.
\newblock In {\em Proceedings of the IEEE conference on computer vision and
  pattern recognition}, pages 5079--5087, 2015.

\bibitem{vaswani2017Transformer}
Ashish Vaswani, Noam Shazeer, Niki Parmar, Jakob Uszkoreit, Llion Jones,
  Aidan~N. Gomez, \L{}ukasz Kaiser, and Illia Polosukhin.
\newblock Attention is all you need.
\newblock In {\em Proceedings of the 31st International Conference on Neural
  Information Processing Systems}, pages 6000–--6010, 2017.

\bibitem{wang2016cnn}
Jiang Wang, Yi~Yang, Junhua Mao, Zhiheng Huang, Chang Huang, and Wei Xu.
\newblock Cnn-rnn: A unified framework for multi-label image classification.
\newblock In {\em Proceedings of the IEEE conference on computer vision and
  pattern recognition}, pages 2285--2294, 2016.

\bibitem{wang2017JRL}
Jingya Wang, Xiatian Zhu, Shaogang Gong, and Wei Li.
\newblock Attribute recognition by joint recurrent learning of context and
  correlation.
\newblock In {\em Proceedings of the IEEE International Conference on Computer
  Vision}, pages 531--540, 2017.

\bibitem{wang2017RNNPAR}
Jingya Wang, Xiatian Zhu, Shaogang Gong, and Wei Li.
\newblock Attribute recognition by joint recurrent learning of context and
  correlation.
\newblock In {\em Proceedings of the IEEE International Conference on Computer
  Vision}, pages 531--540, 2017.

\bibitem{wang2023MMPTMs}
Xiao Wang, Guangyao Chen, Guangwu Qian, Pengcheng Gao, Xiao-Yong Wei, Yaowei
  Wang, Yonghong Tian, and Wen Gao.
\newblock Large-scale multi-modal pre-trained models: A comprehensive survey.
\newblock {\em Machine Intelligence Research}, 20(4):447--482, 2023.

\bibitem{wang2024pedestrian}
Xiao Wang, Jiandong Jin, Chenglong Li, Jin Tang, Cheng Zhang, and Wei Wang.
\newblock Pedestrian attribute recognition via clip based prompt
  vision-language fusion.
\newblock {\em IEEE Transactions on Circuits and Systems for Video Technology},
  2024.

\bibitem{wang2024empiricalmamba}
Xiao Wang, Weizhe Kong, Jiandong Jin, Shiao Wang, Ruichong Gao, Qingchuan Ma,
  Chenglong Li, and Jin Tang.
\newblock An empirical study of mamba-based pedestrian attribute recognition,
  2024.

\bibitem{wang2021TNL2K}
Xiao Wang, Xiujun Shu, Zhipeng Zhang, Bo~Jiang, Yaowei Wang, Yonghong Tian, and
  Feng Wu.
\newblock Towards more flexible and accurate object tracking with natural
  language: Algorithms and benchmark.
\newblock In {\em Proceedings of the IEEE/CVF Conference on Computer Vision and
  Pattern Recognition}, pages 13763--13773, 2021.

\bibitem{wang2024HDETrack}
Xiao Wang, Shiao Wang, Chuanming Tang, Lin Zhu, Bo~Jiang, Yonghong Tian, and
  Jin Tang.
\newblock Event stream-based visual object tracking: A high-resolution
  benchmark dataset and a novel baseline.
\newblock In {\em Proceedings of the IEEE/CVF Conference on Computer Vision and
  Pattern Recognition}, pages 19248--19257, 2024.

\bibitem{wang2022PARsurvey}
Xiao Wang, Shaofei Zheng, Rui Yang, Aihua Zheng, Zhe Chen, Jin Tang, and Bin
  Luo.
\newblock Pedestrian attribute recognition: A survey.
\newblock {\em Pattern Recognition}, 121:108220, 2022.

\bibitem{wei2024event}
Wenjie Wei, Malu Zhang, Jilin Zhang, Ammar Belatreche, Jibin Wu, Zijing Xu,
  Xuerui Qiu, Hong Chen, Yang Yang, and Haizhou Li.
\newblock Event-driven learning for spiking neural networks.
\newblock {\em arXiv preprint arXiv:2403.00270}, 2024.

\bibitem{wu2021progressive}
Jibin Wu, Chenglin Xu, Xiao Han, Daquan Zhou, Malu Zhang, Haizhou Li, and
  Kay~Chen Tan.
\newblock Progressive tandem learning for pattern recognition with deep spiking
  neural networks.
\newblock {\em IEEE Transactions on Pattern Analysis and Machine Intelligence},
  44(11):7824--7840, 2021.

\bibitem{2022iaacaps}
Junyi Wu, Yan Huang, Zhipeng Gao, Yating Hong, Jianqiang Zhao, and Xinsheng Du.
\newblock Inter-attribute awareness for pedestrian attribute recognition.
\newblock {\em Pattern Recognition}, 131:108865, 2022.

\bibitem{xiang2024spiking}
Shuiying Xiang, Tao Zhang, Shuqing Jiang, Yanan Han, Yahui Zhang, Xingxing Guo,
  Licun Yu, Yuechun Shi, and Yue Hao.
\newblock Spiking siamfc++: Deep spiking neural network for object tracking.
\newblock {\em Nonlinear Dynamics}, 112(10):8417--8429, 2024.

\bibitem{yang2021knowledge}
Jing Yang, Brais Martinez, Adrian Bulat, Georgios Tzimiropoulos, et~al.
\newblock Knowledge distillation via softmax regression representation
  learning.
\newblock International Conference on Learning Representations (ICLR), 2021.

\bibitem{yang2021cascaded}
Yang Yang, Zichang Tan, Prayag Tiwari, Hari~Mohan Pandey, Jun Wan, Zhen Lei,
  Guodong Guo, and Stan~Z Li.
\newblock Cascaded split-and-aggregate learning with feature recombination for
  pedestrian attribute recognition.
\newblock {\em International Journal of Computer Vision}, 129(10):2731--2744,
  2021.

\bibitem{yang2022focal}
Zhendong Yang, Zhe Li, Xiaohu Jiang, Yuan Gong, Zehuan Yuan, Danpei Zhao, and
  Chun Yuan.
\newblock Focal and global knowledge distillation for detectors.
\newblock In {\em Proceedings of the IEEE/CVF Conference on Computer Vision and
  Pattern Recognition}, pages 4643--4652, 2022.

\bibitem{yang2022masked}
Zhendong Yang, Zhe Li, Mingqi Shao, Dachuan Shi, Zehuan Yuan, and Chun Yuan.
\newblock Masked generative distillation.
\newblock In {\em European Conference on Computer Vision}, pages 53--69.
  Springer, 2022.

\bibitem{yang2022vitkd}
Zhendong Yang, Zhe Li, Ailing Zeng, Zexian Li, Chun Yuan, and Yu~Li.
\newblock Vitkd: Practical guidelines for vit feature knowledge distillation.
\newblock {\em arXiv preprint arXiv:2209.02432}, 2022.

\bibitem{yao2024spike}
Man Yao, Jiakui Hu, Zhaokun Zhou, Li~Yuan, Yonghong Tian, Bo~Xu, and Guoqi Li.
\newblock Spike-driven transformer.
\newblock {\em Advances in neural information processing systems}, 36, 2024.

\bibitem{zagoruyko2016paying}
Sergey Zagoruyko and Nikos Komodakis.
\newblock Paying more attention to attention: Improving the performance of
  convolutional neural networks via attention transfer.
\newblock {\em arXiv preprint arXiv:1612.03928}, 2016.

\bibitem{zeng2023braincog}
Yi~Zeng, Dongcheng Zhao, Feifei Zhao, Guobin Shen, Yiting Dong, Enmeng Lu, Qian
  Zhang, Yinqian Sun, Qian Liang, Yuxuan Zhao, et~al.
\newblock Braincog: A spiking neural network based, brain-inspired cognitive
  intelligence engine for brain-inspired ai and brain simulation.
\newblock {\em Patterns}, 4(8), 2023.

\bibitem{zhang2014panda}
Ning Zhang, Manohar Paluri, Marc'Aurelio Ranzato, Trevor Darrell, and Lubomir
  Bourdev.
\newblock Panda: Pose aligned networks for deep attribute modeling.
\newblock In {\em Proceedings of the IEEE conference on computer vision and
  pattern recognition}, pages 1637--1644, 2014.

\bibitem{zhao2023transformerVLT}
Haojie Zhao, Xiao Wang, Dong Wang, Huchuan Lu, and Xiang Ruan.
\newblock Transformer vision-language tracking via proxy token guided
  cross-modal fusion.
\newblock {\em Pattern Recognition Letters}, 168:10--16, 2023.

\bibitem{ZHENG2023105708}
Aihua Zheng, Huimin Wang, Jiaxiang Wang, Huaibo Huang, Ran He, and Amir
  Hussain.
\newblock Diverse features discovery transformer for pedestrian attribute
  recognition.
\newblock {\em Engineering Applications of Artificial Intelligence},
  119:105708, 2023.

\bibitem{zheng2021going}
Hanle Zheng, Yujie Wu, Lei Deng, Yifan Hu, and Guoqi Li.
\newblock Going deeper with directly-trained larger spiking neural networks.
\newblock In {\em Proceedings of the AAAI conference on artificial
  intelligence}, volume~35, pages 11062--11070, 2021.

\bibitem{zhou2023spikingformer}
Chenlin Zhou, Liutao Yu, Zhaokun Zhou, Zhengyu Ma, Han Zhang, Huihui Zhou, and
  Yonghong Tian.
\newblock Spikingformer: Spike-driven residual learning for transformer-based
  spiking neural network.
\newblock {\em arXiv preprint arXiv:2304.11954}, 2023.

\bibitem{zhou2021rethinking}
Helong Zhou, Liangchen Song, Jiajie Chen, Ye~Zhou, Guoli Wang, Junsong Yuan,
  and Qian Zhang.
\newblock Rethinking soft labels for knowledge distillation: A bias-variance
  tradeoff perspective.
\newblock {\em arXiv preprint arXiv:2102.00650}, 2021.

\bibitem{zhou2021temporal}
Shibo Zhou, Xiaohua Li, Ying Chen, Sanjeev~T Chandrasekaran, and Arindam
  Sanyal.
\newblock Temporal-coded deep spiking neural network with easy training and
  robust performance.
\newblock In {\em Proceedings of the AAAI conference on artificial
  intelligence}, volume~35, pages 11143--11151, 2021.

\bibitem{zhou2022spikformer}
Zhaokun Zhou, Yuesheng Zhu, Chao He, Yaowei Wang, Shuicheng Yan, Yonghong Tian,
  and Li~Yuan.
\newblock Spikformer: When spiking neural network meets transformer.
\newblock {\em arXiv preprint arXiv:2209.15425}, 2022.

\bibitem{zhu2022event}
Lin Zhu, Xiao Wang, Yi~Chang, Jianing Li, Tiejun Huang, and Yonghong Tian.
\newblock Event-based video reconstruction via potential-assisted spiking
  neural network.
\newblock In {\em Proceedings of the IEEE/CVF Conference on Computer Vision and
  Pattern Recognition}, pages 3594--3604, 2022.

\bibitem{zhu2023spikegpt}
Rui-Jie Zhu, Qihang Zhao, Guoqi Li, and Jason~K Eshraghian.
\newblock Spikegpt: Generative pre-trained language model with spiking neural
  networks.
\newblock {\em arXiv preprint arXiv:2302.13939}, 2023.

\end{thebibliography}
\end{document}